\RequirePackage{fix-cm}
\documentclass[smallextended]{svjour3}       
\smartqed  
\usepackage{graphicx}

\PassOptionsToPackage{warn}{textcomp}  
\usepackage{textcomp}                  
\usepackage{mathptmx}                  
\usepackage{times}                     
\usepackage{tabu}                      
\usepackage{booktabs}                  

\usepackage{amsmath}
\usepackage[ruled,vlined]{algorithm2e} 
\usepackage{subfig}

\begin{document}

\title{Facilitating Machine Learning Model Comparison and Explanation Through A Radial Visualisation
}

\titlerunning{Facilitating Machine Learning Model Comparison}        

\author{Jianlong Zhou         \and
        Weidong Huang \and Fang Chen 
}


\institute{J. Zhou and F. Chen \at
              Data Science Institute\\ 
              University of Technology Sydney, Australia \\
              \email{\{jianlong.zhou, fang.chen\}@uts.edu.au}           
           \and
           W. Huang \at
              Faculty of Transdisciplinary Innovation\\ 
              University of Technology Sydney, Australia \\
              \email{weidong.huang@uts.edu.au} 
}


\maketitle

\begin{abstract}

Building an effective Machine Learning (ML) model for a data set is a difficult task involving various steps. One of the most important steps is to compare generated substantial amounts of ML models to find the optimal one for the deployment. It is challenging to compare such models with dynamic number of features. Comparison is more than just finding differences of ML model performance, users are also interested in the relations between features and model performance such as feature importance for ML explanations. 
This paper proposes \emph{RadialNet Chart}, a novel visualisation approach to compare ML models trained with a different number of features of a given data set while revealing implicit dependent relations. In RadialNet Chart, ML models and features are represented by lines and arcs respectively. These lines are generated effectively using a recursive function. The dependence of ML models with dynamic number of features is encoded into the structure of visualisation, where ML models and their dependent features are directly revealed from related line connections.
ML model performance information is encoded with colour and line width in RadialNet Chart. Together with the structure of visualisation, feature importance can be directly discerned in RadialNet Chart for ML explanations. 

\end{abstract}

\section{Introduction}

\begin{figure*}[!htb]
    \centering 
    \includegraphics[width=0.9\linewidth]{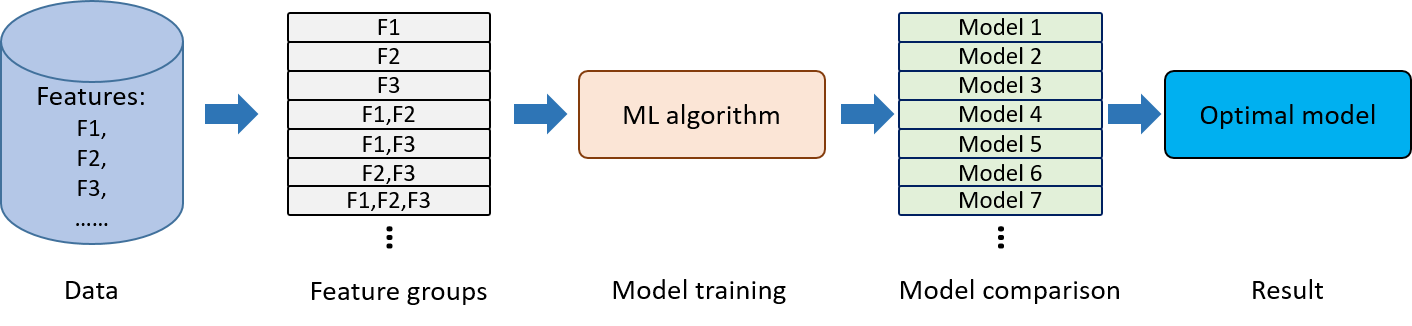}
    \caption{The pipeline of getting an optimal ML model for a data set with multiple features.}
    \label{fig:ml_pipeline}
\end{figure*}

We have witnessed a rapid boom of data in recent years from various fields such as infrastructure, transport, energy, health, education, telecommunications, and finance. Together with the dramatic advances in Machine Learning (ML), getting insights from these ``Big Data'' and data analytics-driven solutions are increasingly in demand for different purposes. While these ``Big Data'' are used by sophisticated ML algorithms to train ML models which are then evaluated by various metrics such as accuracy, the generated substantial amounts of ML models must be compared by the engineering designers and analysts to find the optimal one for the deployment \cite{Zhou_wrapping_2017}. 
Fig.~\ref{fig:ml_pipeline} shows a typical pipeline that processes data to find an optimal ML model. Taking a data set with multiple features for ML training as an example, multiple features can be grouped differently as the input for an ML algorithm to train different ML models. For example, if a data set has three features of F1, F2, and F3, these features may have seven different groups: [F1], [F2], [F3], [F1, F2], [F1, F3], [F2, F3], and [F1, F2, F3]. Each feature group can be used as the input for an ML algorithm to train an ML model, thereby obtaining seven different ML models. It is a common thread to find the best/worst model by comparing such models, however it is often challenging when having a large number of features. 
Furthermore, comparison is more than just finding differences of ML model performance, users are also interested in the relations between features and model performance from comparison to get explanation of models, for example, to find which features result in high performance of ML models, and those features are referred as high important features, or vice versa. This is because the identification of the most or least important features are the key steps for feature engineering in effective and explainable machine learning \cite{Zhou_wrapping_2017,Zhou_human_2018}. 

On the other hand, it is widely recognised that visualisations amplify
human's cognition during data analysis \cite{Card_readings_1999} and proper visualisation of ML outcomes is essential for a human analyst to be able to interpret them \cite{becker_visualizing_2002,talbot_ensemblematrix:_2009}. Viegas and Wattenberg \cite{Viegas_visualization_2017} claimed that ``data visualisation of the performance of algorithms for the purpose of identifying anomalies and generating trust is going to be the major growth area in data visualisation in the coming years''. More importantly, comparison with visualisation is imperative to identify the optimal model from substantial amounts of ML models. Bar chart, radar chart, line chart as well as others \cite{Aigner_vis_2011} are commonly used visualisation methods in machine learning to compare different variables. However, comparison of ML models with a large number of features is still considered challenging with the aid of these commonly used visualisations: the items for comparison and the relationships between them can be highly complicated. While these commonly used visualisation approaches not only cause information clutters for large number of visual elements (e.g. bars, dots, lines) but also miss relation information between features and models, which are significant in ML explanations. It is also very difficult for users to differentiate differences of various model performances with these commonly used visualisation approaches. Despite the specific focus on visualising comparison in recent studies \cite{Gleicher_considerations_2018,Law_duet_2019,Ondov_face_2019}, little work has been done on the visual comparison of ML models while identifying relations between features and ML models (e.g. the most and least important features).

We explore an approach based on the structure of visualisation in addressing challenges of comparison ML models with dynamic number of features: while height information of bars and lines in commonly used visualisation approaches only encode one-dimensional information in a 2-dimensional (2D) space, it is possible to encode ML model information in other dimensions of the space. If both visual elements and structure of visualisation can be used to encode information of ML models, insights about ML models could be automatically generated, users would not have to inspect every model to find optimal one or conduct complex calculations \cite{Zhou_human_2018} to estimate feature importance.


In this paper, we propose \emph{RadialNet Chart} (also referred to RadialNet in this paper), a novel visualisation approach to compare ML models with different number of features while revealing implicit dependent relations. In RadialNet, ML models and features are represented by lines and arcs respectively (an arc also represents the model based on the single feature of arc).  
The challenge of revealing dependence of ML models with dynamic number of features is addressed by encoding such information into the structure of visualisation, where ML models and their dependent features are directly revealed from related line connections. These lines are defined using a recursive function to generate them effectively. 
ML model performance information is encoded with colour and line width in RadialNet.
It simplifies the comparison of different ML models based on these visual encoding. Moreover, together with the structure of visualisation, feature importance can be directly discerned in RadialNet for ML explanation. 
RadialNet uses a concept of feature path for ML model lines to avoid a large number of line entangles. And when visual elements for ML models are crowded, RadialNet allows to interactively change spanning space that RadialNet covers to dynamically control the visual complexities. 
To understand the effectiveness of RadialNet, we conducted a comparison experiment with three commonly used visualisation approaches of line chart, bar chart, and radar chart. The comparison experiment was evaluated with eleven researchers and developers experienced in machine learning related areas. The findings show that RadialNet has advantages in identifying features related to specific models as well as directly revealing importance of features (for ML explanations). 
Furthermore, RadialNet is more efficient to help users focus their attention to find visual elements of interest. 
It is more compact to show more information in a limited space compared with other visualisation types.

\section{Background and Related Work}

In machine learning, given a fixed number of features, it is possible to use different features and their groups to train machine learning algorithms resulting in various machine learning models. 
Users need to compare these models to find the optimal one for their tasks. Getting the optimal results out of machine learning models requires a truly understanding of all models. However,
each data set with a large number of features can have hundreds or even thousands of ML models, making it nearly impossible to understand all models based on different feature groups in an intuitive fashion. Visualisation can be used to help unlock nuances and insights in ML models. 

This section investigates various visualisations from the perspectives of multi-attribute data visualisation, visualisation in explanation of machine learning, and comparison visualisation in order to demonstrate the state-of-art approaches and challenges for comparison of machine learning models with visualisation.

\subsection{Visualisation of Multi-Attribute Data}

The comparison visualisation of machine learning models is related to multi-attribute (or multiple features) data visualisation.
The visualisation of multi-attribute data has been frequently investigated for years.
One of classical approaches to visualise multi-attribute data points is parallel coordinates \cite{Zhou_a_2008}. The advantage of this technique is that it can provide an overview of data trend. One of obvious disadvantages of parallel coordinates is that it lacks a tabular view for presenting value details of each coordinates. SimulSort \cite{Kim_does_2012} organizes different attributes of data in a tabular and sorts all of the attribute columns simultaneously. However, users still need laborious interactions in SimulSort in order to highlight different points for comparison. 
Zhou et al. \cite{zhou_measurable_2015} proposed a visualisation approach for presenting multi-attribute data by combining advantages of both parallel coordinates and SimulSort, which organizes various attributes in a tabular-like form implicitly. Colours are used to encode data belonging to different groups, instead of highlighting attributes of one point at a time as in SimulSort. Such colour encoding approach provides an overview of points and their associated attribute details to improve the information browsing efficiency. Motivated by such colour encoding, this paper uses colours to encode ML model performance to provide an overview of performance for comparison. However, such visualisation cannot reveal complex relations between machine learning models and their dependent features with dynamic numbers.

Moreover, the contradiction between the limited space and the large amount of information to be presented is another challenge for multi-attribute data visualisation.
Coordinated \& multiple views (CMV) \cite{Roberts_state_2007} is widely used to extend the limited space of a single view for large data set visualisation. 
Langner et al. \cite{Langner_vistiles2018} presented a framework that uses a set of mobile devices to distribute and coordinate multiple visualisation views for the exploration of multivariate data.
Koytek et al. \cite{Koytek_mybrush_2018} proposed \emph{MyBrush} for extending brushing and linking technique by incorporating personal agency in the interactive exploration of data relations in CMV.
Sarikaya et al. \cite{Sarikaya_scatterplots_2018} introduced a framework to help determine the design appropriateness of scatterplot for task support to modify/expand the traditional scatterplots to scale as the complexity and amount of data increases. Most of these investigations focus on the extension of spaces for the complex information presentation, however ignore making full use of a given limited space. Our approach in this paper aims to encode complex information with less visual elements (e.g. model lines) to avoid entangled visual elements in the limited space to improve the information presentation efficiency.

\subsection{Visualisation in Explanation of Machine Learning}

In the early years, visualisation is primarily used to explain the ML process of simple ML algorithms in order to understand the ML process. For example, different visualisation methods are used to examine specific values and show probabilities of picked objects visually for Na{\"i}ve-Bayes \cite{becker_visualizing_2002}, decision trees \cite{ankerst_visual_1999}, Support Vector Machines (SVMs) \cite{caragea_gaining_2001}. Advanced visualisation techniques are then proposed to present more complex ML processes. Erra et al. \cite{erra_interactive_2011} introduced a visual clustering which utilises a collective behavioral model, where visualisation helps users to understand and guide the clustering process. Paiva et al. \cite{paiva_improved_2011} presented an approach that employs the similarity tree visualisation to distinguish groups of interest within the data set.
Visualisation is also used as an interaction interface for users in machine learning. For example, Guo et al. \cite{guo_nugget_2011} introduced a visual interface named Nugget Browser allowing users to interactively submit subgroup mining queries for discovering interesting patterns dynamically. EnsembleMatrix allows users to visually ensemble multiple classifiers together and provides a summary visualisation of results of these multiple classifiers \cite{talbot_ensemblematrix:_2009}. Zhou et al. \cite{zhou_making_2016} revealed states of key internal variables of ML models with interactive visualisation to let users perceive what is going on inside a model.

More recent work tries to use visualisation as an interactive tool to facilitate ML diagnosis. ModelTracker \cite{amershi_modeltracker2015} provides an intuitive visualisation interface for ML performance analysis and debugging. Chen et al. \cite{chen_diagnostic_2016} proposed an interactive visualisation tool by combining ten state-of-the-art visualisation methods in ML (shaded confusion matrix, ManiMatrix, learning curve, learning curve of multiple models, McNemar Test matrix, EnsembleMatrix, Customized SmartStripes, Customized ModelTracker, confusion matrix with sub-categories, force-directed graph) to help users interactively carry out a multi-step diagnosis for ML models. Wongsuphasawat et al. \cite{Wong_visualizing_2018} presented the TensorFlow Graph Visualizer to visualise data flow graphs of deep learning models in TensorFlow to help users understand, debug, and share the structure of their deep learning models.
Neto and Paulovich \cite{neto_explainable_2020} presented a matrix-like visual metaphor to explain rule-based Random Forest models, where logic rules are rows, features are columns, and rules predicates are cells. The visual explanation allows users to effectively obtain overviews of models (global explanations) and audit classification results (local explanations). 
Chan et al. \cite{chan_melody_2020} presented an interactive visual analytics system to construct an optimal global overview of the model and data behaviour by summarising the local explanations using information theory. The explanation summary groups similar features and instances from local explanations and is presented with visualisations.

Visualisations comprise the major body of ML process explanations. However, these approaches cannot be directly used for the comparison of machine learning models trained with a different number of features, and facilitate the revealing of feature importance directly from visualisations of models for ML explanations. 

\subsection{Comparison Visualisation}

Supporting comparison is a common challenge in visualisation. Gleicher \cite{Gleicher_considerations_2018} categorized four considerations that abstract comparison when using visualisation. These four considerations include to identify: the comparative elements, the comparative challenges, a comparative strategy, and a comparative design, which provide a guideline for developing comparison solutions in visualisation. 
Law et al. \cite{Law_duet_2019} presented Duet, a visual analysis system to conduct pairwise comparisons. Duet employs minimal specification in comparison by only recommending similar and different attributes between them when one object group to be compared is specified.

Bar chart is one of commonly used visualisation methods for comparison in machine learning \cite{Aigner_vis_2011}.
It works with two variables -- one is the length of the bar on one axis and the second is the position of this bar on another axis. The variable is compared by denoting it with the length of the bars when various bars are plotted together. Radar Chart is another commonly used approach to compare multiple quantitative variables. It is useful for seeing which variables have similar values or if there are any outliers amongst the values of each variable. It can also help to find which variables are high or low. Besides, other methods such as line chart and ring chart are also used in comparison. Ondov et al. \cite{Ondov_face_2019} made evaluations of comparison visualizations of 5 layouts: stacked small multiples, adjacent small multiples, overlaid charts, adjacent small multiples that are mirror symmetric and animated transitions. The data to be compared are encoded with the length of bars in bar charts, slop of lines in line charts, and angle of arcs in donut charts.

These previous works provide significant guidelines and advances in comparison visualisation. This paper proposes a new visualisation method for machine learning model comparison with a full consideration of four aspects as categorized in \cite{Gleicher_considerations_2018}. The new visualisation approach is evaluated by comparing it with other three commonly used visualisation methods (bar chart, line chart, and radar chart) in machine learning model comparisons.


\section{RadialNet Chart}

This section presents a novel visualisation approach called \emph{RadialNet Chart} to compare machine learning models trained with different feature groups of a data set.

\subsection{Design Goals}
After having a thorough survey with experienced researchers and developers in machine learning on their problems meeting in comparing machine learning models, we phrase following design goals for the RadialNet:

\begin{itemize}
  \setlength\itemsep{-0.2em}
  \item \textbf{Comparison}: To maximise differences among visual elements of models to help users find the optimal target easily. The comparison is the core objective in the ML model visualisation. This is a challenge when substantial amounts of ML models must be compared. 
  
  \item \textbf{Importance}: To easily identify importance of features directly from visualisation. The importance of features plays significant roles in the feature selection in the ML pipeline and ML explanations \cite{Biran2017}. It is a challenge to identify importance of features directly from visualisation without complex feature importance calculations.
  
  \item \textbf{Feature identification}: To easily identify relationships between models (and model performance) and their dependent features. This helps users easily link ML models and their dependent features for understanding both features and models, which is usually challenging with commonly used visualisation approaches.
  
  \item \textbf{Compactness}: To represent complex visualisation in a compact form and reduce the visual clutters because of substantial amounts of information in a limited space.
\end{itemize}

\subsection{Definition of RadialNet Chart}

This subsection defines the RadialNet. Fig.~\ref{fig:sc_demo} shows an example of RadialNet. Based on this example, we firstly give following definitions that are used to set up a RadialNet:

\begin{figure}[tb]
    \centering 
    \includegraphics[width=0.5\linewidth]{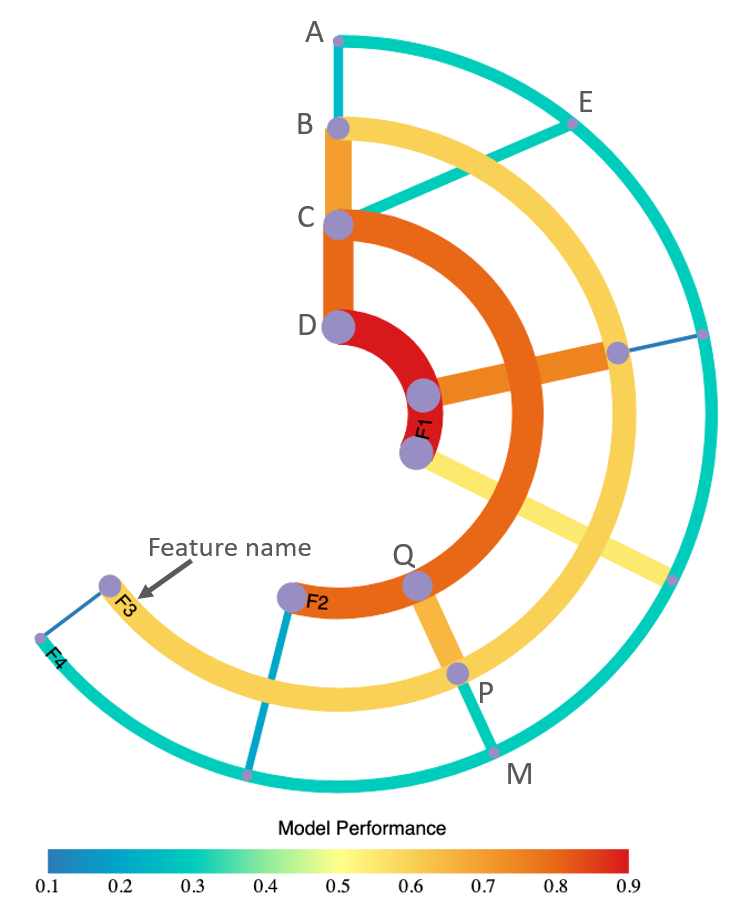}
    \caption{An example of RadialNet chart.}
    \label{fig:sc_demo}
\end{figure}

\begin{description}
    \setlength\itemsep{-0.2em}
  \item[Feature arc] Each feature is represented by a concentric arc in RadialNet. The arc is also called feature arc. The name of each feature is displayed at one end of the arc as shown in Fig.~\ref{fig:sc_demo} (e.g. F1, F2, F3, F4). Each arc also represents the ML model based on that single feature. 
  
  \item[Model line] RadialNet uses a line segment to represent an ML model based on multiple features. The line is also called model line. For example, in Fig.~\ref{fig:sc_demo}, the line $AB$, $BC$, and $CD$ represent different ML models respectively. The features used for the model are defined based on the feature path of the line (see the definition of feature path below).
  
  \item[Feature point] A feature point refers to an intersection point of a model line with an arc. It is represented by a dot point on a feature arc as shown in Fig.~\ref{fig:sc_demo} (e.g. feature points A, B, C).
  
  \item[Feature path]  A feature path defines features used for a model line. A feature path starts from the feature point of a model line on its outermost arc and ends at the feature point on the innermost arc it can reach through the connected feature point in the RadialNet. For example, in Fig.~\ref{fig:sc_demo}, for the model line $AB$, its feature path starts from the feature point A on the arc F4, passes through B and C, and ends at D on the innermost arc F1. This path can be represented by a list of features corresponding to arcs of each feature point, i.e. the feature path of $AB$ is [F4, F3, F2, F1]. Similarly, the feature path of $BC$ is [F3, F2, F1], the feature path of $CD$ is [F2, F1], the feature path of $EC$ is [F4, F2, F1], the feature path of $MP$ is [F4, F3, F2], and the feature path of $PQ$ is [F3, F2].
\end{description}

Furthermore, the model performance is encoded using two methods: the width of the line/arc and the colour of the line/arc. The wider the line/arc is, the higher the model performance. A colour scale is accompanied with the RadialNet to encode model performance and let users easily perceive the difference of performance of different models as shown in Fig.~\ref{fig:sc_demo}.

Based on these definitions, the visualisation of lines and arcs are spiraling from the centre to outside and therefore it is called \emph{RadialNet Chart}. 
The RadialNet has different advantages. For example,
given a data set in machine learning, if most of ML models related to one specific feature show high model performance, that feature can be considered as a high important feature, and vice versa if most of ML models related to one specific feature show low model performance, that feature can be considered as a less important feature. The RadialNet can depict importance of features directly through visualisation: if an arc and its connected lines are mostly wider than others and have colours representing high performance values in the colour scale, the feature represented by the arc is an important feature, and vice versa it can also depict less important features. For example, in Fig.~\ref{fig:sc_demo}, the feature F1 is an important feature because the width and colour of the arc as well as its connected lines are mostly wider and red, while the feature F4 is an less important feature. 
The RadialNet also helps users directly identify features used for a specific model because of the feature path mechanism in RadialNet.

\begin{figure}[htb]
    \centering 
    \includegraphics[width=0.9\linewidth]{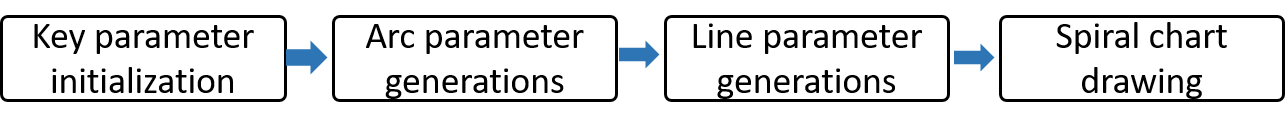}
    \caption{The steps for drawing RadialNet.}
    \label{fig:drawing_pipeline}
\end{figure}

Fig.~\ref{fig:drawing_pipeline} shows the steps used to draw a RadialNet. The definition of different parameters is the key during RadialNet drawing. Firstly, key parameters are defined with user interactions or predefined approaches. Arc parameters and line parameters are then generated based on key parameters. The RadialNet is drawn finally based on generated parameters.

\subsection{Key Parameter Initialization}

The key parameters include the overall spanning angle of RadialNet, the overall number of models given the number of features, the size of the drawing canvas, as well as others. The overall spanning angle defines the space that the RadialNet covers in degrees. It can be interactively modulated by users to control the compactness of the visualisation in a limited space. If the number of ML models to be visualized is low, a small value can be defined for the spanning angle, and vice versa a large value can be defined for the spanning angle in order to help users easily control and compare ML models in a limited space.

Given $N$ features of a data set, F1, F2, ..., FN, a machine learning algorithm uses these features to set up ML models. The ML models can be set up based on one or multiple features of the data set. Typically, the number of models based on various groups of $N$ features can be got from Equ.~\ref{equ:num_models}:

\begin{equation}
    \label{equ:num_models}
    C_N = C_{N}^{1} + C_{N}^{2} + ... + C_{N}^{i} + ... + C_{N}^{N}=2^{N}-1
\end{equation}

\noindent where $C_N$ is the number of models based on groups of $N$ features, $C_{N}^{i}$ is the group number of selecting $i$ features from $N$ features. It shows that the number of ML models is increased exponentially with the increase of number of features.

Furthermore, because of the circular characteristics of RadialNet, polar coordinates are used to represent arcs and lines in RadialNet.

\subsection{Arc Parameter Generations}

Algorithm \ref{algo:arcParaGen} shows the process for generating arc parameters. The arc is denoted by its start point and end point in polar coordinates. In this algorithm, 
\emph{\texttt{arcSpanning}} defines the largest angle that arcs cover in the space and can be interactively changed by a sliding bar in the user interface. \emph{\texttt{N}} is the number of features. \emph{\texttt{canvasWidth}} is the width of the drawing canvas. \emph{\texttt{allFeatures}} is a list of all studied features which are sorted in the decreased order based on model performance of individual features. Each arc represents the model performance based on an individual feature from \emph{\texttt{allFeatures}} list. The algorithm generates arc parameters aiming to make $N$ arcs evenly distributed in the drawing canvas space. This algorithm initialises the spanning angle of each arc with the \emph{\texttt{arcSpanning}} value, and the spanning of each arc (\emph{\texttt{arcAngle}}) is dynamically updated in the drawing algorithm (see Algorithm~\ref{algo:spiralchart}) to allow arcs in a spiral format. \emph{\texttt{arcParasDict}} is a dictionary storing parameters of arcs and the key of the dictionary is the individual features for the arc. The parameters include arc's radius, spanning angle and arc width. \emph{\texttt{Data}} is read from a JSON file and stores different feature groups and their model performance values.

\begin{algorithm}
\caption{Algorithm for arc parameter generations}
\label{algo:arcParaGen}
\LinesNumbered
\SetKwProg{Fn}{Function}{:}{end}
\SetKwFunction{FArcParasGen}{\textsc{ArcParasGen}}
\Fn{ \FArcParasGen{ \texttt{arcSpanning, N, canvasWidth, allFeatures, Data}}} {
    \tcp{Distance between two arcs}
    arcSpacing $\gets$ \emph{\texttt{canvasWidth}}/(2*\emph{\texttt{N}})\;
    prev\_radius $\gets$ 0.0\;
    \emph{\texttt{arcParasDict}} $\gets$ \{ \}\;
    \For{f {\upshape in} \texttt{allFeatures}}{
        arcRadius $\gets$ prev\_radius + arcSpacing\;
        prev\_radius $\gets$ arcRadius\;
        \tcp{Encode performance of the model based on f as the arc width}
        arcWidth $\gets$ \emph{\texttt{Data}}[$f$].performance\;
        arcAngle $\gets$ \emph{\texttt{arcSpanning}}\;
        \emph{\texttt{arcParasDict}}[$f$] $\gets$ [arcRadius, arcAngle, arcWidth]\;
    }
   \Return \emph{\texttt{arcParasDict}}
}  
\end{algorithm}

\subsection{Line Parameter Generations}

Algorithm \ref{algo:lineParaGen} shows a recursive function used for generating model line parameters. The line is denoted by its start point and end point in polar coordinates. In this algorithm, \emph{\texttt{lineParasDict}} is a dictionary and stores parameters of lines, and the key of the dictionary is the feature list (feature path) used for the line. The line parameters stored in the dictionary include the start and end points of the line in polar coordinates as well as line width of the line.
\emph{\texttt{lineFeatures}} is the feature list for the current line and is sorted in the decreased order based on model performance of individual features. \emph{\texttt{startAngle}} is the angle of polar coordinates of the start point of the line. \emph{\texttt{angleStep}} is the step size that angle increases each time.

In this algorithm, if the key with the current \emph{\texttt{lineFeatures}} does not exist in \emph{\texttt{lineParasDict}}, a sub-key with the feature list by removing the last feature in \emph{\texttt{lineFeatures}} is created. If this sub-key still does not exit in \emph{\texttt{lineParasDict}} and the number of features in this sub-key is more than 2, the algorithm recursively call this function with the current sub-key features. Otherwise, the algorithm defines the start point and end point of the line and pushes them into \emph{\texttt{lineParasDict}}.

The line width is encoded with the model performance based on \emph{\texttt{lineFeatures}}. The colour of the line is also encoded with the model performance using a colour scale.

\begin{algorithm}[!htb]
\caption{Algorithm for line parameter generations}
\label{algo:lineParaGen}
\LinesNumbered
\SetKwProg{Fn}{Function}{:}{end}
\SetKwFunction{FLinePara}{\textsc{LineParasGen}}
\SetKw{Not}{not}
\SetKw{And}{and}
\SetKw{In}{in}
\SetKwData{AllFeatures}{allFeature}\SetKwData{LineFeatures}{lineFeatures}
\SetKwData{lineParasDict}{lineParasDict}\SetKwData{arcParasDict}{arcParasDict}
\SetKwData{Ikey}{ikey}\SetKwData{Isubkey}{isubkey}
\Fn{ \FLinePara{\texttt{allFeatures, lineParasDict, arcParasDict, lineFeatures, startAngle, angleStep, Data}}}{
  \tcp{Use lineFeatures as key of lineParasDict}
  $ikey$ $\gets$ \emph{\texttt{lineFeatures}}\;
  len\_lineFeatures $\gets$ \emph{\texttt{lineFeatures}}.length\;
  \If {ikey {\upshape is} \Not \In \texttt{lineParasDict}} {
    \tcp{Sub-features without the last feature}
      $isubkey$ $\gets$ $ikey$[:len\_lineFeatures-1]\;
      len\_isubkey $\gets$ $isubkey$.length\;
      \uIf {isubkey {\upshape is} \Not \In \texttt{lineParasDict} \And {\upshape len\_isubkey}$>2$} {
        \tcp{Recursively call the function}
        \FLinePara {\texttt{allFeatures, lineParasDict, arcParasDict}, isubkey, \texttt{startAngle, angleStep, Data}}\;}
      \Else{
        \tcp{Define start and end points}
        \uIf {isubkey {\upshape is} \In \texttt{lineParasDict}} {
            \tcp{Polar coordinates of start point}
            \emph{\texttt{startAngle}}$\gets$\emph{\texttt{lineParasDict}}[\emph{isubkey}].endAngle\;
            startRadius$\gets$\emph{\texttt{lineParasDict}}[\emph{isubkey}].endRadius\;
            endSubF $\gets$ \emph{isubkey}.endFeature\;
            endF $\gets$ \emph{ikey}.endFeature\;
            \If {\Not neighbour{\upshape(endSubF, endF)} \In \texttt{allFeatures}}{
                dist $\gets$ distance(endF, endSubF) \In \emph{\texttt{allFeatures}}\;
                \emph{\texttt{startAngle}}$\gets$\emph{\texttt{startAngle}}+\emph{\texttt{angleStep}}*dist\;
            }
        }
        \Else{
            \If{\texttt{lineFeatures}.{\upshape length} == 2}{ 
                \emph{\texttt{startAngle}}$\gets$\emph{\texttt{startAngle}}+\emph{\texttt{angleStep}}\;
           }
           iFeature $\gets$ \emph{\texttt{lineFeatures}}[len\_lineFeatures-1]\;
           startRadius $\gets$ \emph{\texttt{arcParasDict}}[iFeature].radius\;
        }
        \tcp{Polar coordinates of end point}
        lastFeature $\gets$ \emph{\texttt{lineFeatures}}[len\_lineFeatures]\;
        endAngle $\gets$ \emph{\texttt{startAngle}}\;
        endRadius $\gets$ \emph{\texttt{arcParasDict}}[lastFeature]\;
        \tcp{Encode model performance as the line width}
        lineWidth $\gets$ \emph{\texttt{Data}}[\emph{\texttt{lineFeatures}}].performance\;
        \tcp{Push line parameters into dict}
        \emph{\texttt{lineParasDict}}[\emph{ikey}] $\gets$ [\emph{\texttt{startAngle}}, startRadius, endAngle, endRadius, lineWidth]\;
      }
   }
   
 
   \Return{\texttt{lineParasDict, startAngle}}
}  
\end{algorithm}

\subsection{RadialNet Chart Drawing}

Algorithm \ref{algo:spiralchart} shows the process of drawing a RadialNet. In Algorithm~\ref{algo:spiralchart}, after getting key parameters such as number of points on the outermost arc and arc spanning angle, Algorithm~\ref{algo:arcParaGen} is firstly called to generate arc parameters. Then Algorithm~\ref{algo:lineParaGen} is called for each feature to generate line parameters related to that feature. These parameters are then used to draw arcs and lines by calling functions of DrawArcs() and DrawLines() respectively. DrawArcs() and DrawLines() calls Javascript functions to draw arcs and lines.

\begin{algorithm}[!htb]
\caption{Algorithm for drawing RadialNet}
\label{algo:spiralchart}
\LinesNumbered
 \KwIn{\emph{\texttt{allFeatures, arcSpanning, N, canvasWidth, Data}}}
 \KwOut{SpiralChart}
 \SetKw{In}{in}
 \SetKw{To}{to}
 \BlankLine
\tcp{Number of points on the outmost arc}
num\_points $\gets C_{N-1}$; \tcp*[h]{see Equ.~\ref{equ:num_models}}\; 
\tcp{Define step size of angles}
angleStep $\gets$ 2*\emph{\texttt{arcSpanning}} / (num\_points - 1)\;
\tcp{Initialize parameters}
\emph{\texttt{startAngle}} $\gets$ 0\;
\emph{\texttt{lineParasDict}} $\gets$ \{ \}\;
\BlankLine
\tcp{Generate arc parameters}
\emph{\texttt{arcParasDict}} $\gets$ \textsc{ArcParasGen}(\emph{\texttt{arcSpanning, N, canvasWidth}})\;
\BlankLine
\tcp{Generate line parameters}
\For{ $f$ \In \texttt{allFeatures }}{
    \tcp{Number of lines based on feature {\upshape f}}
    num\_lines $\gets$ \emph{\texttt{Data}}[$f$].length\;
    \BlankLine
    \For{$j \gets 1$ \To {\upshape num\_lines}}{
        \tcp{Feature list used for the current line}
        \emph{\texttt{lineFeatures}} $\gets$ \emph{\texttt{Data}}[$f$][$j$]\;
        \tcp{Number of features for the current line}
        num\_features $\gets$ \emph{\texttt{lineFeatures}}.length\;
        \If {{\upshape num\_features} != {\upshape 1}} {
            \tcp{Generate line parameters}
            \emph{\texttt{lineParasDict, startAngle}} $\gets$ \textsc{LineParasGen} (\emph{\texttt{allFeatures, lineParasDict, arcParasDict, lineFeatures, startAngle, angleStep}})\;
            \BlankLine
            \tcp{Update arcAngle}
            \If{j == {\upshape num\_lines}}{
                \emph{\texttt{arcParasDict}}[$f$].arcAngle $\gets$ \emph{\texttt{startAngle}}/2\;
            }
        }
    }
 }
 \BlankLine
 \tcp{DrawLines and DrawArcs call Javascript functions to draw lines and arcs of RadialNet Chart}
 DrawArcs (\emph{\texttt{arcParasDict}})\;
 DrawLines (\emph{\texttt{lineParasDict}})\;
\end{algorithm}

\section{Implementation}

The proposed approach is implemented in Javascript based on the D3.js library \cite{Bostock_d3_2011}. The data input to RadialNet are saved in a JSON file. The RadialNet is also implemented as a Javascript library and it is easily to be reused in different visualisation applications. This library will be released as an open-source library.

\section{Case Studies}

\begin{figure}[htb]
    \centering 
    \includegraphics[width=0.55\linewidth]{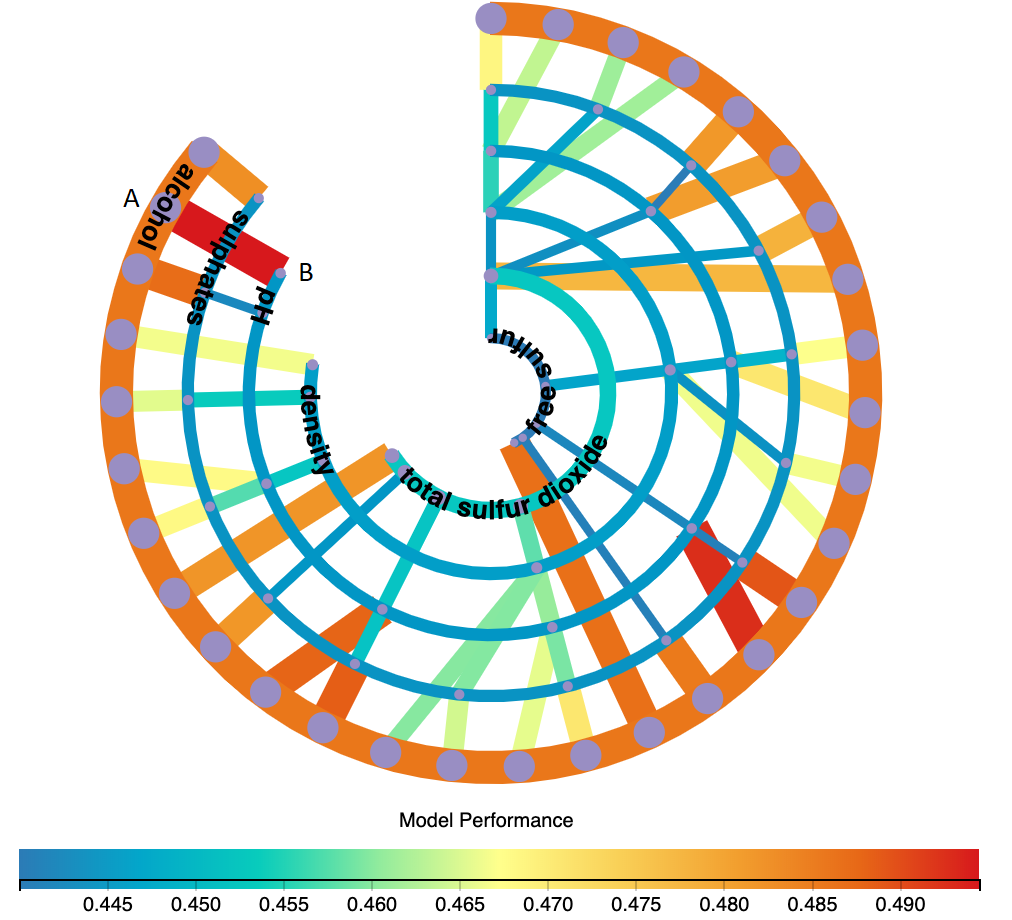}
    \caption{RadialNet of ML models based on a data set with 6 features.}
    \label{fig:spiralchart_small}
\end{figure}

\begin{figure}[htb]
    \centering 
    \includegraphics[width=0.5\linewidth]{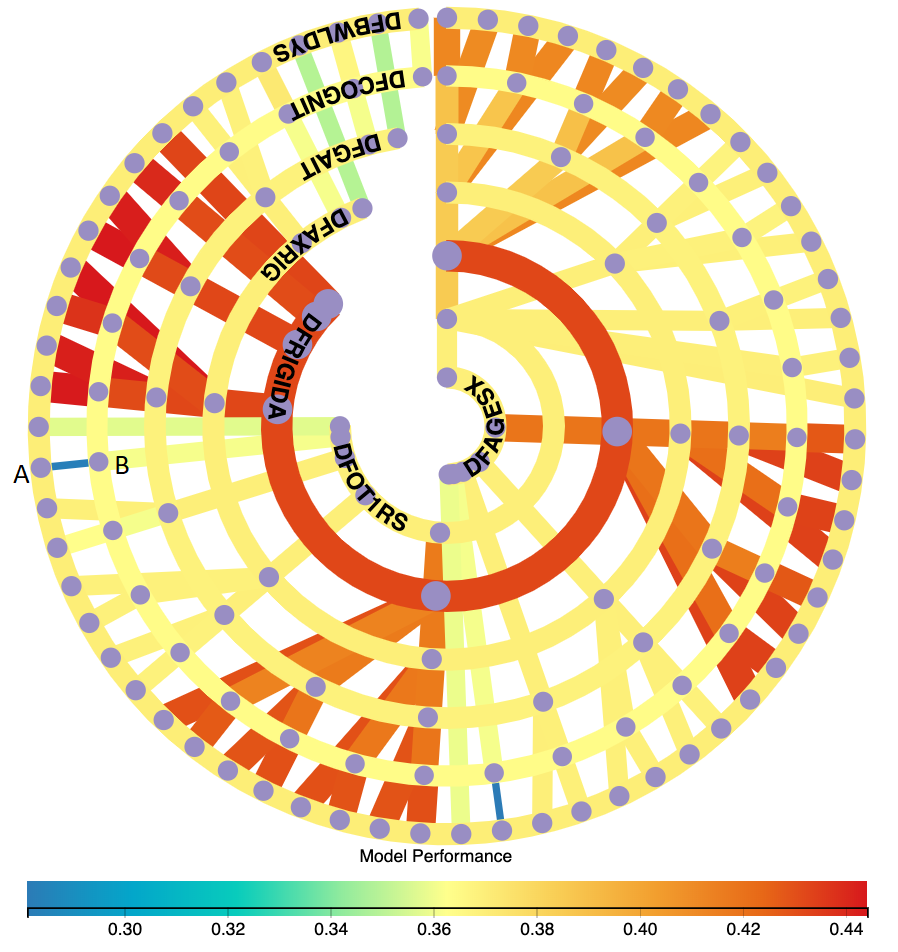}
    \caption{RadialNet of ML models based on a data set with 7 features.}
    \label{fig:spiralchart_big}
\end{figure}

In this section, RadialNet is used to visualise machine learning models based on different data sets and ML algorithms. Two data sets from UCI machine learning data repository \cite{Dua_2017} and PPMI \cite{Prakash_feasibility_2019} respectively were analyzed, and three machine learning algorithms of K-Nearest Neighbours (KNN), Na{\"i}ve Bayes (NB) and Random Forest (RF) were deployed in the experiment. Fig.~\ref{fig:spiralchart_small} shows the visualisation of different ML models for a data set with 6 features. From this visualisation, we can easily locate the model with the highest performance (the widest red line $AB$ as shown in Fig.~\ref{fig:spiralchart_small}) as well as features (two features of ``alcohol'' and ``pH'' on the feature path of the line) used for the model training. It also helps users easily identify the importance of features, the most important feature ``alcohol'' is represented by the outermost arc (the arc and its connected lines are mostly redder and wider than others) and the lest important feature ``free suffur'' is represented by the innermost arc (the arc and its connected lines are mostly bluer and narrower than others). 
Fig.~\ref{fig:spiralchart_big} shows the visualisation of different ML models for a data set with 7 features. Compared with Fig.~\ref{fig:spiralchart_small}, the model number is increased dramatically when the feature number is increased just one. This visualisation also helps users easily locate the model with the lowest performance (the narrowest blue line $AB$ as shown in Fig.~\ref{fig:spiralchart_big}). We can also easily directly identify the most important feature (the third inner arc represented by the widest red arc) and the least important feature (the innermost narrowest yellow arc) as shown in Fig.~\ref{fig:spiralchart_big}.

\begin{figure*}[!htb]
    \centering 
    \includegraphics[width=0.82\linewidth]{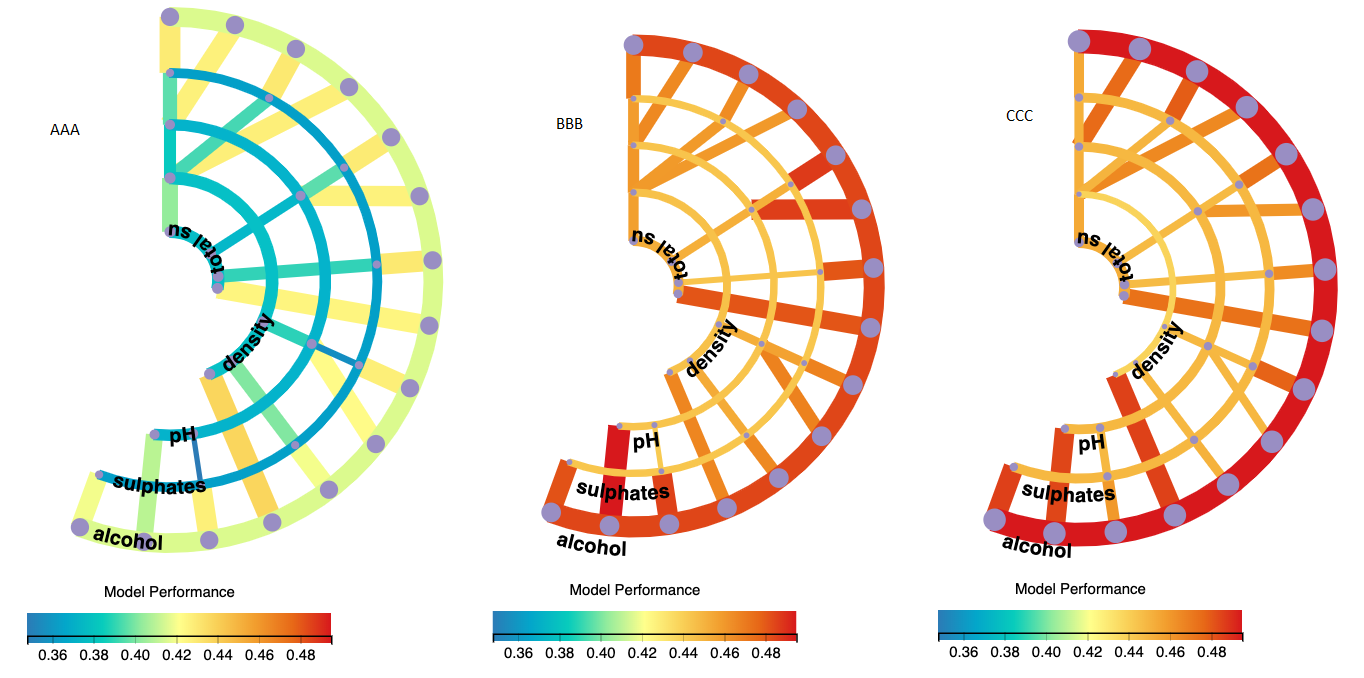}
    \caption{Comparison of three ML algorithms for the same data set with RadialNet.}
    \label{fig:spiralchart_comparison}
\end{figure*}

Besides comparison of feature importance of a data in RadialNet, it can also be used to compare performance of different ML algorithms for a given data set. Fig.~\ref{fig:spiralchart_comparison} shows the comparison of three ML algorithms for the same data set with RadialNet visualisation. From this figure, we can easily get that the ML algorithm represented by the left diagram shows the worst performance, compared to algorithms represented by the other two diagrams, because its colour is bluer which is located on the left side of the colour scale. While the algorithm represented by the middle diagram shows the best performance because its colour is redder which is located on the right side of the colour scale. Furthermore, the visualisation shows that the feature represented by the outermost arc (i.e. the feature of ``alcohol'') is the most important feature because this arc is the widest and its colour is located on the right side of the colour scale in all three visualizations.

\section{Evaluation}

To understand the effectiveness of RadialNet in the ML model comparison, we compare it with three commonly used visualisation approaches of bar chart, line chart and radar chart. 
11 participants were recruited (9 males and 2 females, ages from 20s-40s) to conduct a comparison user study. All participants are researchers and developers experienced in machine learning related areas. 

The following metrics are proposed to evaluate different visualisations:
\begin{itemize}
    \setlength\itemsep{-0.2em}
    \item \textbf{Comparison}: How easily that the visualisation helps users to compare performance of different models;
    \item \textbf{Feature importance}: How easily that the visualisation helps users to identify importance of features;
    \item \textbf{Feature identification}: How easily that the visualisation helps users to link each model and its dependent features;
    \item \textbf{Complexity}: How complex the visualisation is to present data.
\end{itemize}

Besides, user cognitive responses to visualisation such as mental effort as well as time spent on the selection task are also evaluated to compare effectiveness of visualizations:
\begin{itemize}
    \setlength\itemsep{-0.2em}
    \item \textbf{Mental effort}: How much mental effort users used for tasks with the visualisation;
    \item \textbf{Time spent}: How much time users spent in task decisions with the visualisation.
\end{itemize}

To understand the usability of the RadialNet Chart, we also administrate a questionnaire that asks participants questions about their experience and feedback in using the charts. Further, eye tracking study is conducted with a separate participant to understand participant's eye movement behaviours with different visualisations.

\subsection{Data and Visualisation}

\begin{figure*}[t!]
    \centering
    \subfloat[Bar chart]{
        \centering
        \includegraphics[width=0.42\linewidth]{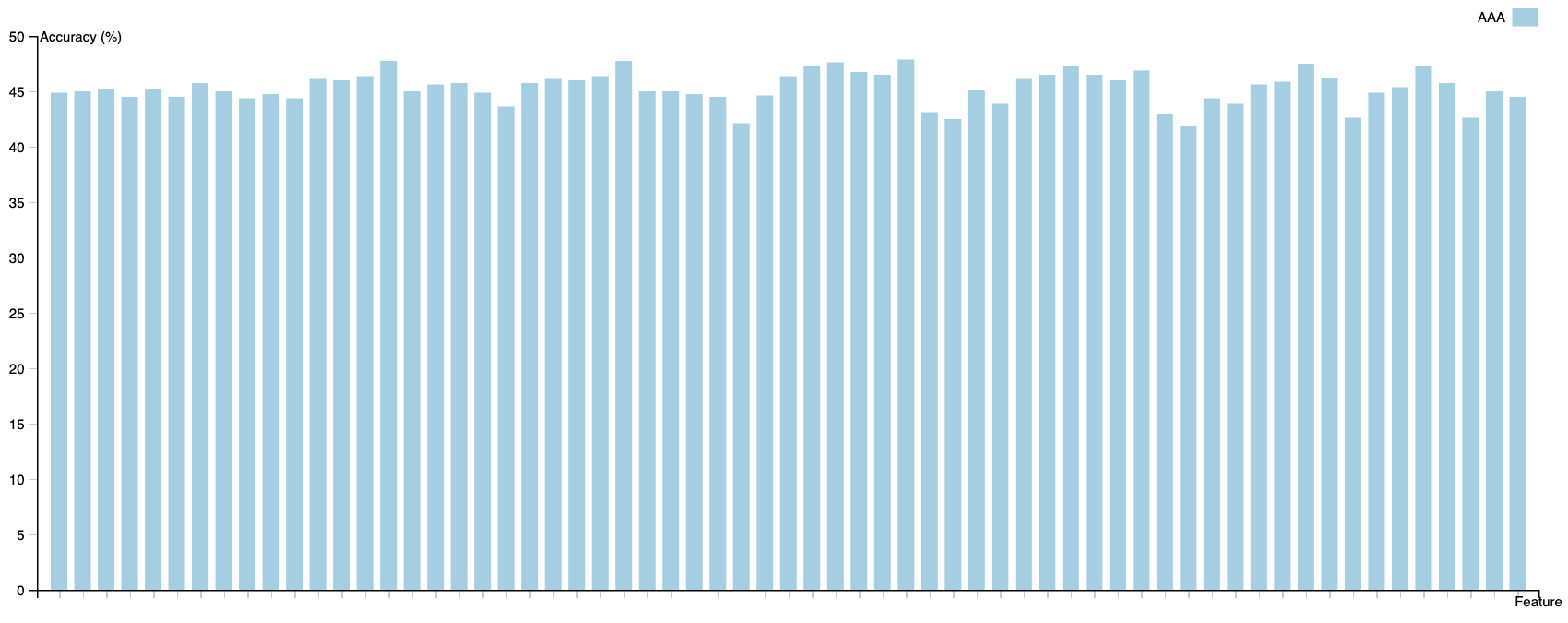}
        \label{fig:barchat_small_sig}
    }
    \subfloat[Line chart]{
        \centering
        \includegraphics[width=0.38\linewidth]{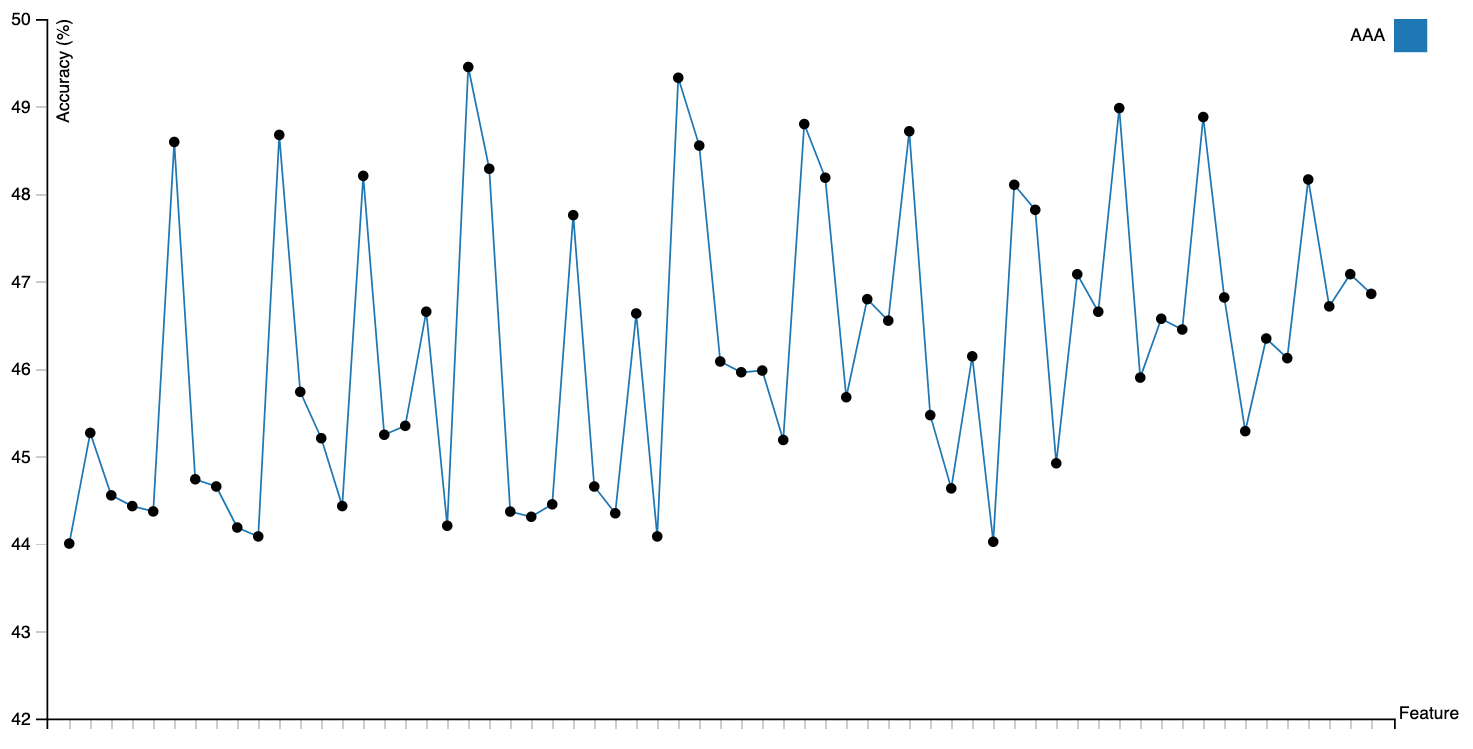}
        \label{fig:linechart_small_sig}
    }
    \subfloat[Radar chart]{
        \centering
        \includegraphics[width=0.18\linewidth]{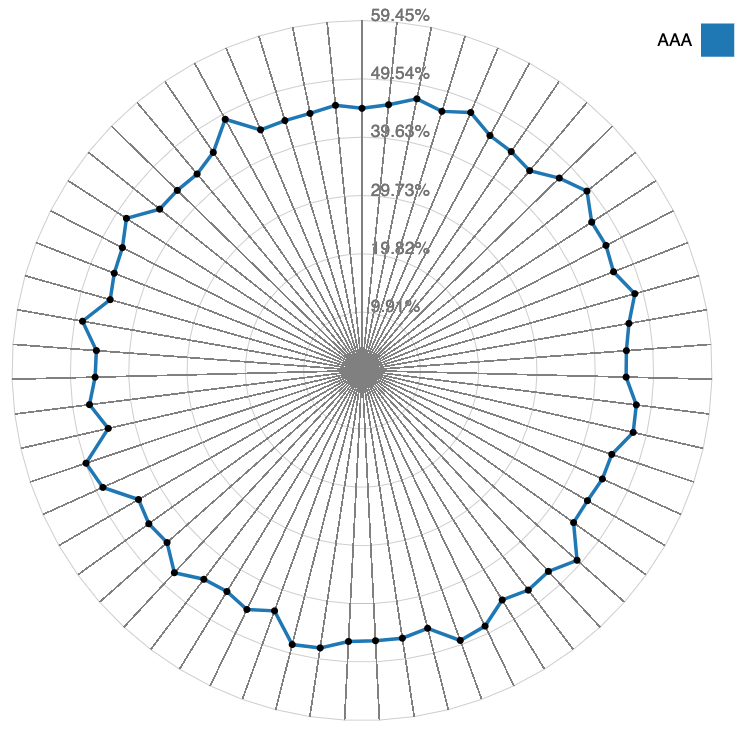}
        \label{fig:radarchart_small_sig}
    }
    \caption{ML models based on the data set of 6 features are visualized using bar chart, line chart and radar chart respectively.}
    \label{fig:threecharts_small}
\end{figure*}

Two data sets from UCI machine learning data repository \cite{Dua_2017} and PPMI \cite{Prakash_feasibility_2019} respectively were analysed in this study. Two data sets have 6 features and 7 features respectively, which generate 63 ML models and 127 ML models respectively to compare. ML models are visualised using bar chart, line chart, radar chart, and RadialNet respectively as shown in Fig.~\ref{fig:threecharts_small} and Fig.~\ref{fig:spiralchart_small} (the data set with 6 features visualised in Fig.~\ref{fig:threecharts_small} and Fig.~\ref{fig:spiralchart_small}). In bar chart, line chart, radar chart and RadialNet, the related features for a model and its performance are popped up when the mouse is moved over the relevant visual elements (e.g. bars, dots, lines, or arcs), which allows users to inspect more details of each model.

\begin{figure*}[t!]
    \centering
    \subfloat[Bar chart]{
        \centering
        \includegraphics[width=0.42\linewidth]{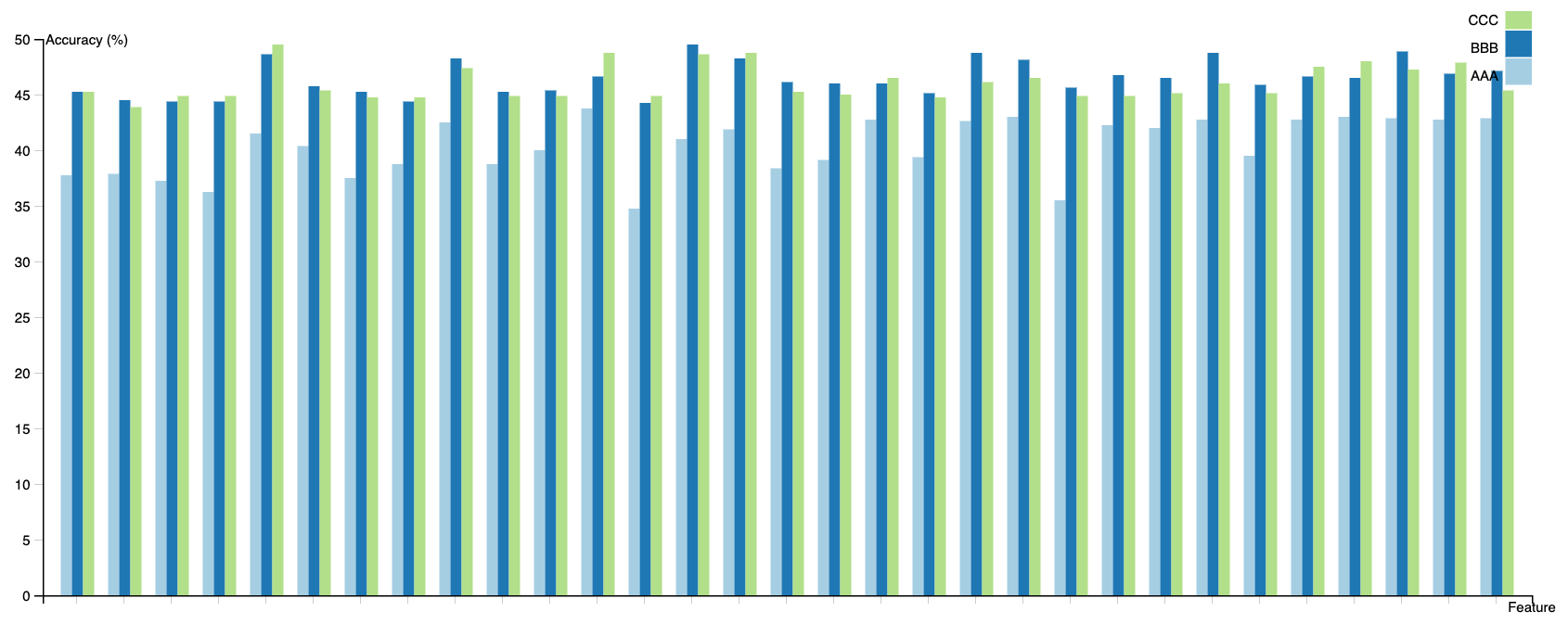}
        \label{fig:barchart_small_comp}
    }%
    \subfloat[Line chart]{
        \centering
        \includegraphics[width=0.38\linewidth]{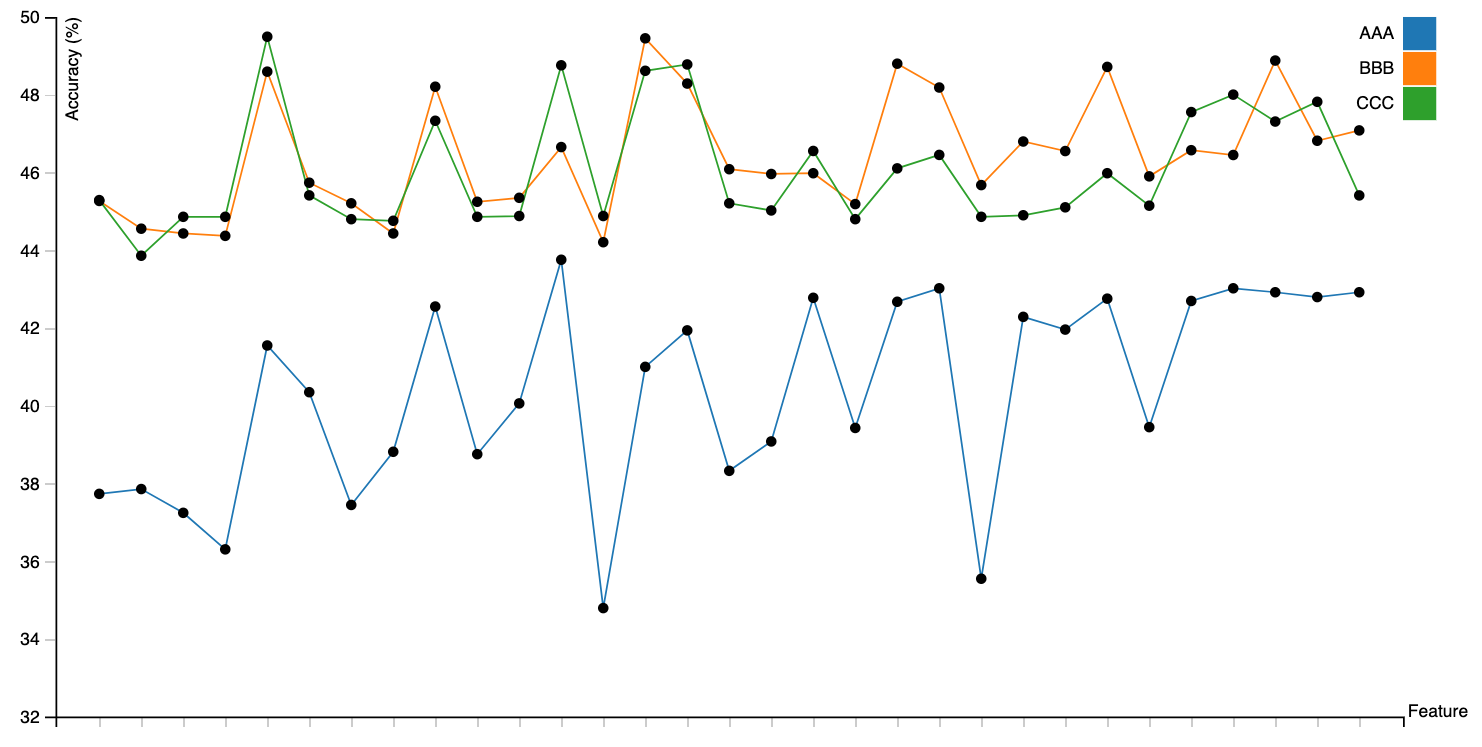}
        \label{fig:linechart_small_comp}
    }%
    \subfloat[Radar chart]{
        \centering
        \includegraphics[width=0.18\linewidth]{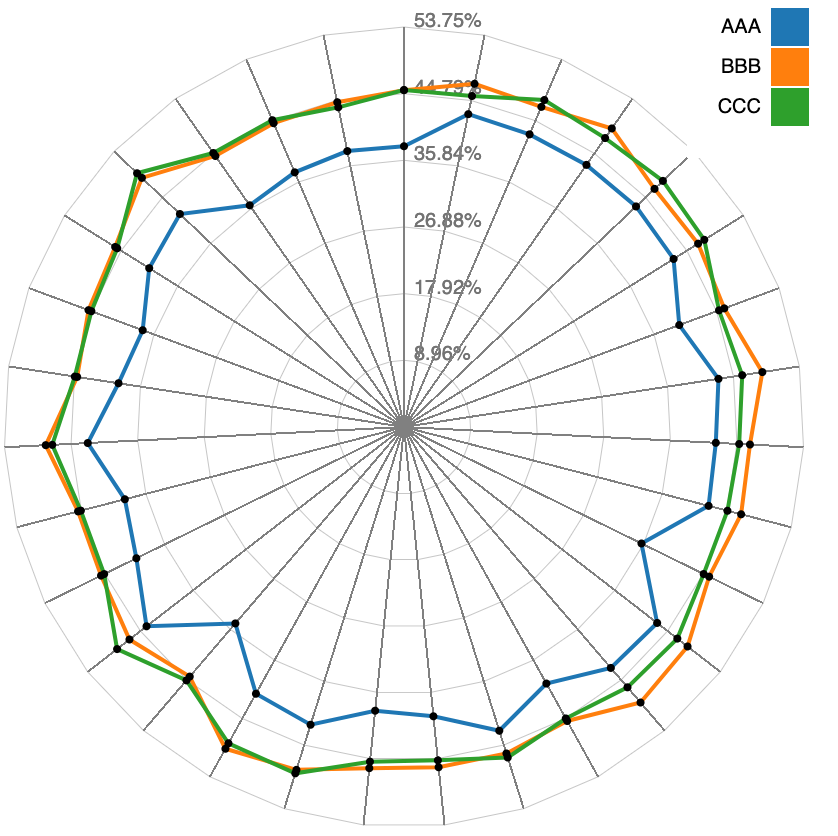}
        \label{fig:radarchart_small_comp}
    }
    \caption{Comparison of three ML algorithms for the same data set with three visualisation approaches.}
    \label{fig:threecharts_small_comp}
\end{figure*}

Besides, for a given data set, three ML algorithms were used generating various ML models respectively. The ML models by these three ML algorithms were visualised together in a single bar chart, line chart, and radar chart respectively as shown in Fig.~\ref{fig:threecharts_small_comp}, which were also visualised using RadialNet as shown in Fig.~\ref{fig:spiralchart_comparison}. These visualisations were used to compare the effectiveness of different ML algorithms.
AAA, BBB, and CCC in visualisations (e.g. Fig.~\ref{fig:threecharts_small}, Fig.~\ref{fig:threecharts_small_comp}) represent three ML algorithms used to compare: KNN, NB, and RF. The exact ML algorithms used for ML models were not shown to participants during the study to avoid any bias.

\subsection{Procedure and Data Collection}

The study was conducted in a lab environment using a Macbook Pro with 13-inch display of resolution $2560\times1600$. 
The procedure of the study is described as follows: Tutorial slides on the study were firstly presented to participants to let them understand concepts and operations during the study. A training task was then conducted to practice interactions. After that, the formal tasks were conducted with different visualisations. During the study, different visualisations as described in the previous section were displayed to participants one-by-one in random order. For each visualisation, participants were firstly required to find which ML model gives the best or worst performance by selecting the visual elements in the visualisation (we call it as the selection task). This is more akin to what analysts do with real data sets. After the selection task, participants were asked to answer different questions as described below on the task and visualisation. At the end of the study, participants were asked to give their feedback in using the charts and some personal details such as gender, age, working topics.

After the selection task of each visualisation, the participants were asked to answer questions related to comparison, feature importance, feature identification, visual complexity, and mental effort on the visualisation using 9-point Likert scales (comparison, feature importance, feature identification: 1=least easiness, 9=most easiness; visual complexity: 1=least complex, 9=most complex; mental effort: 1=least effort, 9=most effort). At the end of all visualisation tasks, the participants were also asked to answer in a questionnaire which visualisation helps users more easily compare ML performance of different features, and which visualisation helps users more easily compare ML performance of different ML algorithms respectively.

\subsection{Results}

In this section, for the evaluation of each metrics, we firstly performed one-way ANOVA test and then followed it up with post-hoc analysis using t-tests (with a Bonferroni correction under a significance level set at $p< \frac{.05}{4}=.013$, based on the fact that we had four visualisation types to test) to analyze differences in participant responses of each metrics. Each metric values were normalised with respect to each subject to minimise individual differences in rating behavior (see Equ.~\ref{equ:metric_norm}): 

\begin{equation}
\label{equ:metric_norm} 
T_i^N=\frac{T_i-T_i^{min}}{T_i^{max}-T_i^{min}}             
\end{equation}

\noindent where $T_i$ and  $T_i^N$ are the original metric rating and the normalised metric rating respectively from the participant $i$, $T_i^{min}$ and $T_i^{max}$ are the minimum and maximum of metric ratings respectively from the participant $i$ in all of his/her tasks. The time spent in the selection tasks is also normalised in a similar way as other five metrics.

\begin{figure*}[t!]
    \centering
    \subfloat[Comparison easiness]{
        \centering
        \includegraphics[width=0.33\linewidth]{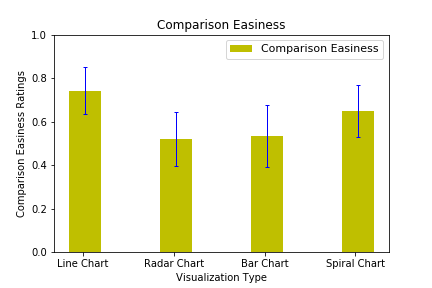}
        \label{fig:Comparison_Easiness_sig_test}
    }%
    \subfloat[Feature identification]{
        \centering
        \includegraphics[width=0.33\linewidth]{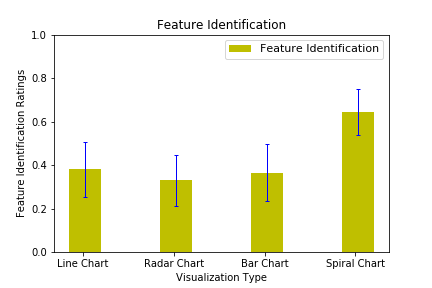}
        \label{fig:Feature_Identification_sig_test}
    }%
    \subfloat[Feature importance]{
        \centering
        \includegraphics[width=0.33\linewidth]{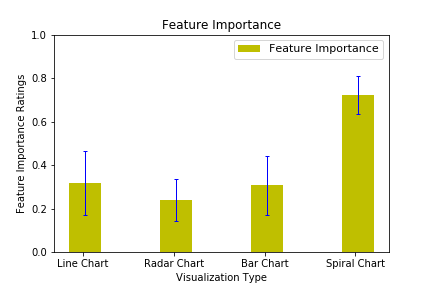}
        \label{fig:Feature_Importance_sig_test}
    }\\
    \subfloat[Visual complexity]{
        \centering
        \includegraphics[width=0.33\linewidth]{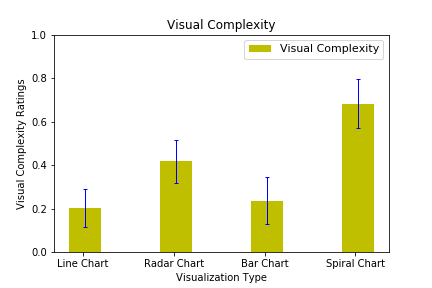}
        \label{fig:Visual_Complexity_sig_test}
    }
    \subfloat[Mental effort]{
        \centering
        \includegraphics[width=0.33\linewidth]{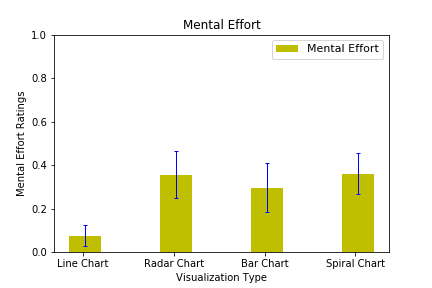}
        \label{fig:Mental_Effort_sig_test}
    }%
    \subfloat[Time spent]{
        \centering
        \includegraphics[width=0.33\linewidth]{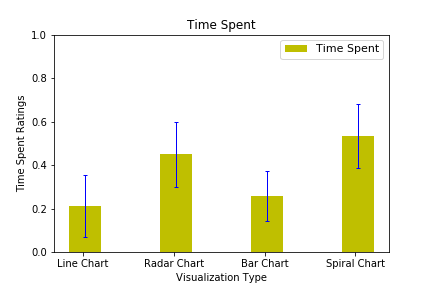}
        \label{fig:Time_Spent_sig_test}
    }%
    \caption{Comparison of mean normalized metrics for different visualisation types.}
    \label{fig:vistype_sig_test}
\end{figure*}

Fig.~\ref{fig:vistype_sig_test} shows mean normalised metric values for different visualisation types.

\begin{description}
  \setlength\itemsep{-0.2em}
  \item[Comparison easiness] One-way ANOVA test gave significant differences in comparison easiness among four visualisation types ($F\left(3,84\right)=3.067$, $p < .03$) (see Fig.~\ref{fig:Comparison_Easiness_sig_test}). However, the post-hoc t-tests only found that line chart was significantly easier to compare performance of different ML models than radar chart ($t = 2.813, p < .007$). The result shows that RadialNet did not help users increase the easiness in comparing performance of different ML models, which is not as we expected, but a trend shows the higher ratings in comparison easiness for RadialNet than bar chart and radar chart (see Fig.~\ref{fig:Comparison_Easiness_sig_test}). This is maybe because of the relatively small number of participants used for the study.
  
  \item[Feature identification] One-way ANOVA test found significant differences in easiness of feature identification among four visualisation types ($F\left(3,84\right)=6.108$, $p < .001$) (see Fig.~\ref{fig:Feature_Identification_sig_test}). The post-hoc t-tests found that RadialNet was significantly easier to identify features related to models than all of other three visualisation types (line chart: $t = 3.296, p < .002$; bar chart: $t = 3.393, p < .002$; radar chart: $t = 4.089, p = .000$). This is because that users can get features and performance related to an ML model directly from connected visual elements in RadialNet, while users need to move mouses to visual elements of each model to inspect related features and performance in other three visualisations.
  
  \item[Feature importance] There were significant differences found in easiness of identifying feature importance among four visualisation types by one-way ANOVA test ($F\left(3,84\right)=14.481$, $p = .000$) (see Fig.~\ref{fig:Feature_Importance_sig_test}). The post-hoc t-tests found that RadialNet was significantly easier to identify feature importance than all of other three visualisation types (line chart: $t = 4.878, p = .000$; bar chart: $t = 5.320, p = .000$; radar chart: $t = 7.678, p = .000$). The results suggest the obvious advantage of RadialNet over other three visualisation types for feature importance identifications.
  
  \item[Visual complexity] One-way ANOVA test found significant differences in visual complexity among four visualisation types ($F\left(3,84\right)=20.254$, $p = .000$) (see Fig.~\ref{fig:Visual_Complexity_sig_test}). The post-hoc t-tests found that RadialNet was significantly more complex than all of other three visualisation types (line chart: $t = 7.032, p = .000$; bar chart: $t = 6.001, p = .000$; radar chart: $t = 3.710, p < .001$). It was also found that radar chart was significantly more complex than line chart ($t = 3.383, p < .002$).
  
  \item[Mental effort]  There were significant differences found in mental effort among four visualisation types by one-way ANOVA test ($F\left(3,84\right)=8.757$, $p = .000$) (see Fig.~\ref{fig:Mental_Effort_sig_test}). The post-hoc tests found that line chart took significantly less effort than other three visualisation types (bar chart: $t = 3.722, p < .001$; radar chart: $t = 4.981, p = .000$; RadialNet: $t = 5.562, p = .000$). RadialNet did not show significant differences in mental effort with radar chart and bar chart.
  
  \item[Time spent] One-way ANOVA test found significant differences in time spent in the selection of the best/worst model task among four visualisation types ($F \left(3,84\right)=5.301$, $p < .002$) (see Fig.~\ref{fig:Time_Spent_sig_test}). The post-hoc tests found that users spent significantly more time in RadialNet than in both line chart ($t = 3.286, p < .002$) and bar chart ($t = 3.111, p < .003$) respectively.  
\end{description}

When four types of visualisation were used to compare performance of different ML algorithms for a given data set, it was found that line chart was easier to compare performance of different ML algorithms followed by RadialNet despite no significant differences found in the easiness. This could be because of the relatively small number of participants in this study. However, RadialNet can reveal importance of features while others not when comparing performance of different ML algorithms. 

\begin{figure*}[t!]
    \centering
    \subfloat[Bar chart]{
        \centering
        \includegraphics[width=0.42\linewidth]{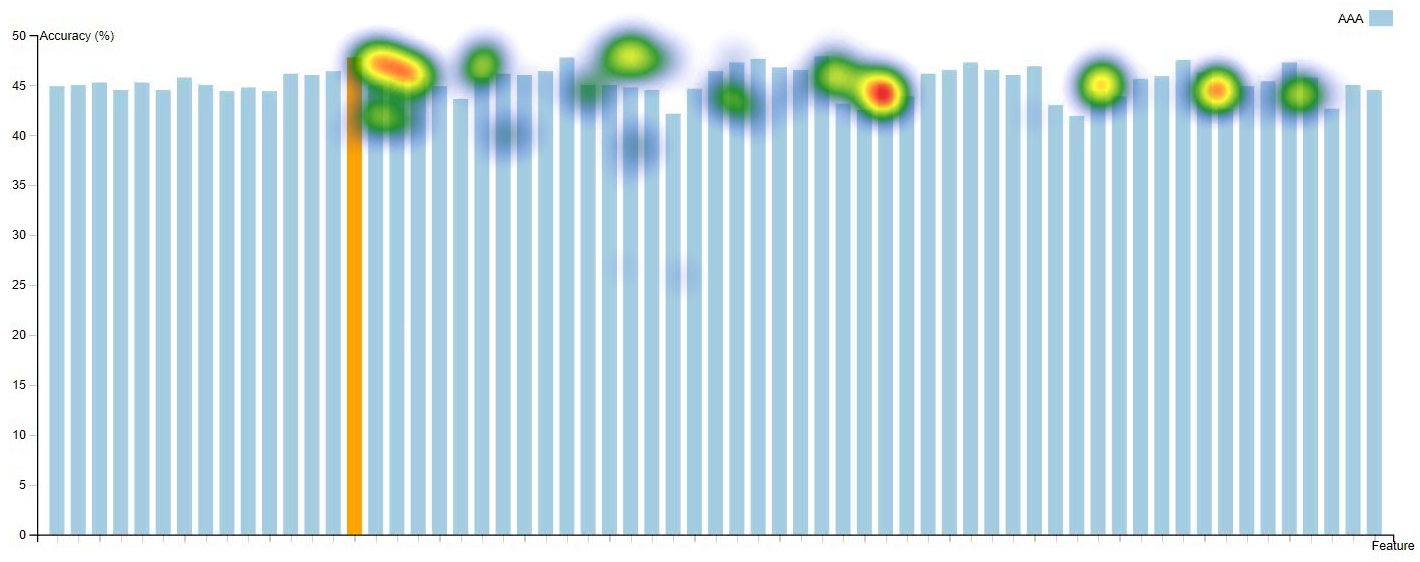}
        \label{fig:barchart_heatmap}
    }%
    \subfloat[Line chart]{
        \centering
        \includegraphics[width=0.38\linewidth]{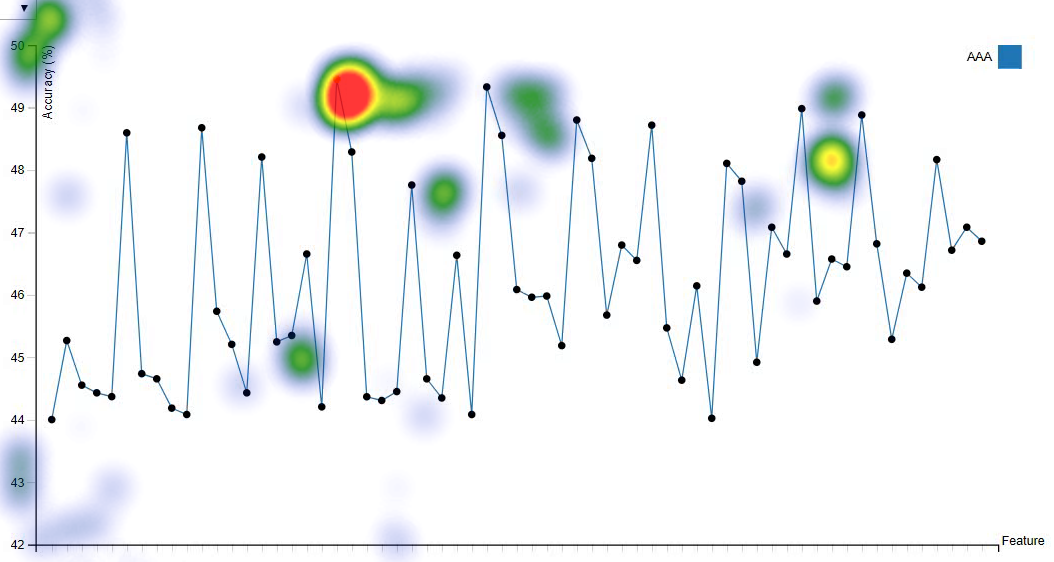}
        \label{fig:linechart_heatmap}
    }%
    \subfloat[Radar chart]{
        \centering
        \includegraphics[width=0.19\linewidth]{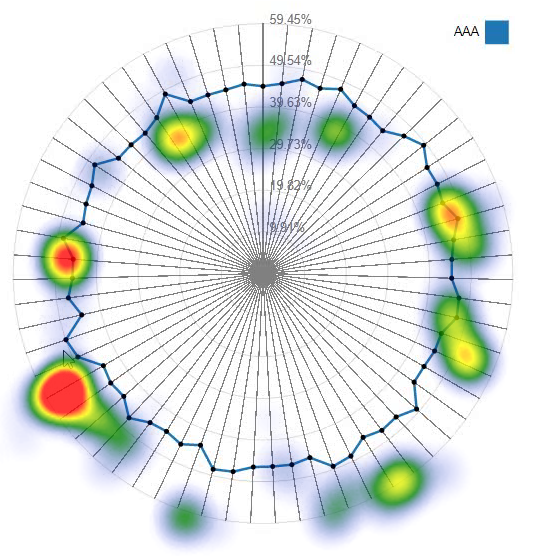}
        \label{fig:radarchart_heatmap}
    }
    \caption{Heat maps of bar chart, line chart, and radar chart.}
    \label{fig:threecharts_heatmap}
\end{figure*}

\begin{figure}[!htb]
    \centering 
    \includegraphics[width=0.5\linewidth]{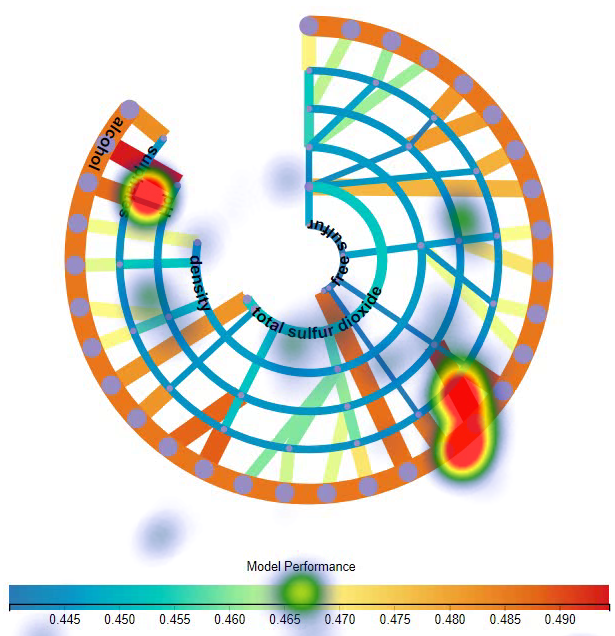}
    \caption{Heat map of RadialNet.}
    \label{fig:spiralchart_heatmap}
\end{figure}

\begin{figure}[!htb]
    \centering 
    \includegraphics[width=0.9\linewidth]{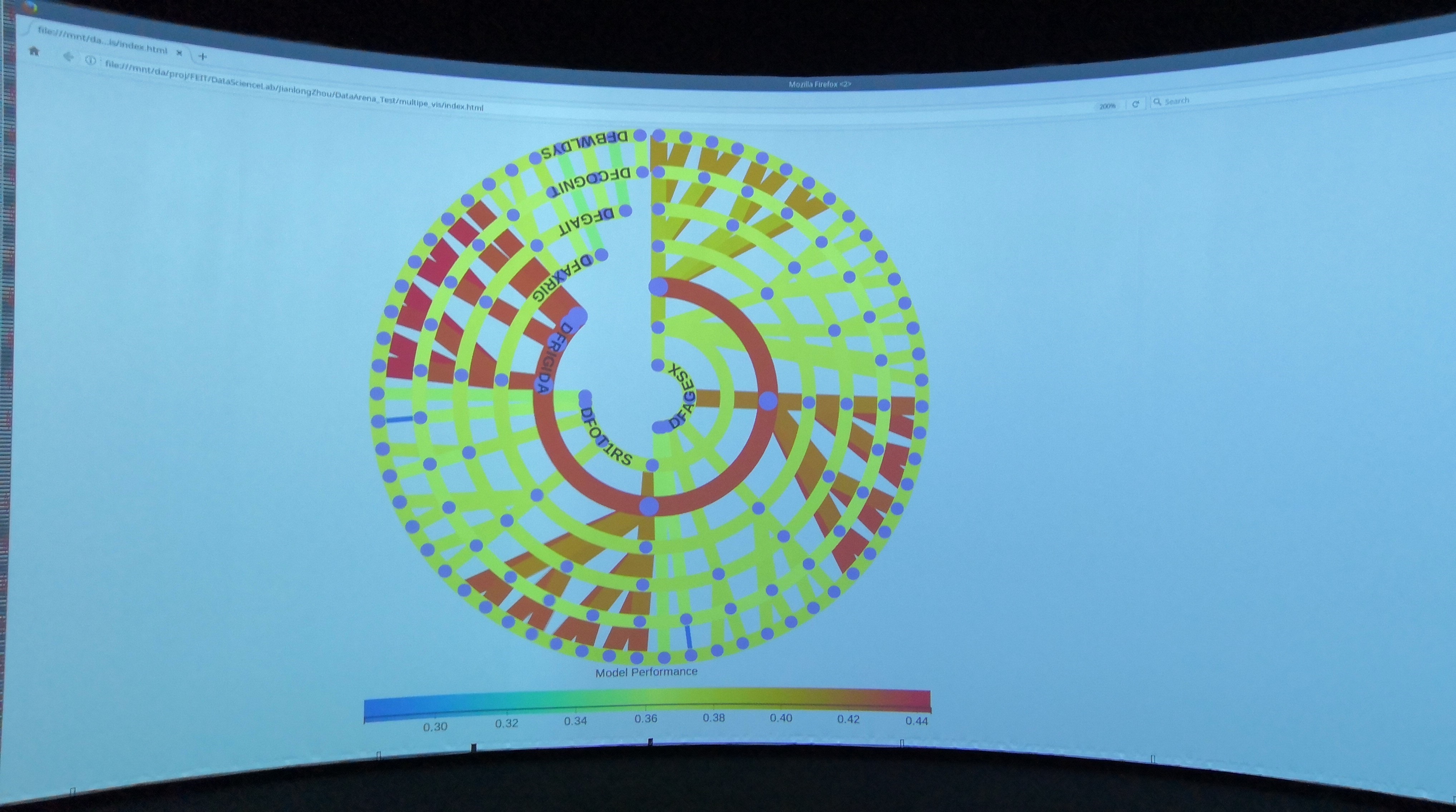}
    \caption{RadialNet displayed in our large scale visualisation facility.}
    \label{fig:spiralchart_dataarena}
\end{figure}

We also collected participants' feedback after completing all tasks by each participant. Overall, all participants believed that ``RadialNet is the most effective visualisation in identifying feature importance compared with other three approaches''. Some participants suggested to ``enlarge the size of RadialNet with the increase of number of features''. 
Participants agreed that ``RadialNet is more efficient to help users focus their attention to find visual elements of interest''. 
Fig.~\ref{fig:threecharts_heatmap} and Fig.~\ref{fig:spiralchart_heatmap} show heat maps on four visualisations recorded by an SMI eye-tracker from a participant during the selection task period respectively. Heat maps reveal the focus of attention by colours indicating the amount of time eyes stay focused on a particular area in the visualisation, the redder, the more time eyes focused.
Fig.~\ref{fig:threecharts_heatmap} and Fig.~\ref{fig:spiralchart_heatmap} suggest that the user's attention in RadialNet was more focused on two model lines with high performance (wide red lines), while it was much scattered among different points in other three visualisations. 

Overall, we can say that RadialNet shows significant advantages in identifying features and performance related to specific models as well as easily revealing importance of features compared with other three visualisation types. Despite these advantages, the mental effort and time spent in RadialNet did not show much differences from others such as radar chart. 

\section{Discussion}

This study proposed a novel visualisation approach to compare variables with different number of dependents. 
Data information is encoded with colour, line width, as well as structure of visualisation to reveal insights from data. 
The experimental results showed that RadialNet has advantages in identifying features related to specific models as well as directly revealing importance of features for ML explanations. 
Different from conventional feature importance evaluations based on complex computing algorithms \cite{Zhou_human_2018} (such as by simulating lack of knowledge about the values of the feature(s) \cite{robnik-sikonja_quality_2012}, or by mean decrease impurity, which is defined as the total decrease in node impurity averaged over all trees of the ensemble in Random Forest\cite{Breiman_classification_1984}), RadialNet allows users to estimate feature importance directly from visualisation by checking lines connected to the feature arc. The consistent large line width of these lines with colours on the right-hand side of the colour scale indicate the high importance of the feature to the modelling.

RadialNet is more compact to show more information in a limited space compared with other three visualisation types. And the compactness of RadialNet can also be controlled by changing its spanning angle dynamically (see the attached video with this paper). However, RadialNet will be much complex when the number of features is high. This could be compensated with large scale visualisation facilities. For example, we have a 360-degree interactive data visualisation facility set to change the way we view and interact with data. Viewers stand in the middle of a large cylindrical screen, four metres high and ten metres in diameter. A high performance computer graphics system drives six 3D-stereo video projectors, edge-blended to create a seamless three-dimensional panorama. Picture clarity is made possible from an image that's $20,000 \times 1200$ pixels. This facility can be used to present RadialNet with large number of ML models for effective interactions. Fig.~\ref{fig:spiralchart_dataarena} shows an example of RadialNet displayed with around 60-degree field of view in the facility.

This paper used the exploration of performance of ML models based on different feature groups from a given data set as a case study to demonstrate the powerfulness of RadialNet in visualising data with complex relations. The RadialNet can also be generalised to other applications where similar relations need to be explored.

\section{Conclusion and Future Work}

This paper presented \emph{RadialNet Chart}, a novel visualisation approach to
compare ML models with different number of features while revealing
implicit dependent relations. The RadialNet is developed to address the challenges faced in comparing a large amount of ML models with each dependent on a dynamic number of features. It is implemented by representing ML models and features with lines and arcs respectively, which in turn are generated by a recursive function and a feature path concept.
We presented our design criteria and described the algorithms for generating the chart. Two case studies were also presented with representative data sets and an experiment was conducted evaluating the effectiveness of the RadialNet. Our case studies showed that the proposed visualisation can help users easily locate target models and important features. Furthermore, the user study revealed that in comparison with other commonly used visualisation approaches, RadialNet is more
efficient to help users focus their attention to find visual elements of
interest. It is also more compact to show more information in a limited
space. Our research provides an effective visualisation approach to
represent data with complex relations. It is specifically helpful for users
to find optimal machine learning model and discern feature importance
visually and directly, but not through complex algorithmic calculations for ML explanations.

For future work, we plan to conduct a more comprehensive eye tracking study, focusing on the analysis of eye activities during the
task time to learn eye movement patterns in viewing RadialNet. Such
study will help to improve the design of RadialNet for more efficient
information browsing. We also hope that the use of RadialNet in
ML model comparison presented here can serve as both a template and a
motivation for other data and applications.
RadialNet Chart will be available as an open-source library.

\bibliographystyle{spmpsci}
\bibliography{chi2019_influ,tml,vis}

\end{document}